\documentclass[sigconf, screen]{acmart}

\usepackage{xspace}
\usepackage{xcolor}
\usepackage{mathtools}
\usepackage{tabularray}
\usepackage{pgfplots}
\usepackage{pgfgantt}
\usepackage{cleveref}
\usepackage{orcidlink}

\newcommand{\biomrc}{{\scshape{biomrc}}\xspace}
\newcommand{\biomrclite}{{\scshape{biomrc lite}}\xspace}
\newcommand{\biomrctiny}{{\scshape{biomrc tiny}}\xspace}
\newcommand{\pubmed}{PubMed\xspace}
\newcommand{\scibert}{Sci\-BERT\xspace}
\newcommand{\biobert}{Bio\-BERT\xspace}
\newcommand{\aoa}{AoA Reader\xspace}
\DeclarePairedDelimiter{\set}{\{}{\}}

\DeclareMathOperator{\softmax}{softmax}
\DeclareMathOperator*{\fa}{F_1}
\DeclareMathOperator*{\fb}{F_2}
\UseTblrLibrary{booktabs}

\renewcommand\footnotetextcopyrightpermission[1]{} %
\begin{document}

\title[Contextual embedding and model weighting on Biomedical Question Answering]{Contextual embedding and model weighting by fusing domain knowledge on Biomedical Question Answering}

\author{Yuxuan Lu\orcidlink{0000-0002-8520-0540}}

\affiliation{%
  \institution{Beijing University of Technology}
  \streetaddress{100 Pingleyuan}
    \city{Chaoyang Qu}
    \state{Beijing Shi}
    \country{China}}
\email{luyuxuanleo@gmail.com}

\author{Jingya Yan\orcidlink{0000-0003-0373-1696}}

\affiliation{%
  \institution{Beijing University of Technology}
  \streetaddress{100 Pingleyuan}
    \city{Chaoyang Qu}
    \state{Beijing Shi}
    \country{China}}
\email{yanjy1998@163.com}

\author{Zhixuan Qi\orcidlink{0000-0001-5333-0335}}

\affiliation{%
  \institution{Beijing University of Technology}
  \streetaddress{100 Pingleyuan}
    \city{Chaoyang Qu}
    \state{Beijing Shi}
    \country{China}}
\email{zhixuanqi@outlook.com}

\author{Zhongzheng Ge\orcidlink{0000-0003-2002-6251}}

\affiliation{%
  \institution{Beijing University of Technology}
  \streetaddress{100 Pingleyuan}
    \city{Chaoyang Qu}
    \state{Beijing Shi}
    \country{China}}
\email{bubble327670@gmail.com}

\author{Yongping Du\orcidlink{0000-0001-6867-2063}}

\affiliation{%
  \institution{Beijing University of Technology}
  \streetaddress{100 Pingleyuan}
    \city{Chaoyang Qu}
    \state{Beijing Shi}
    \country{China}}
\email{ypdu@bjut.edu.cn}

\renewcommand{\shortauthors}{Yuxuan Lu, et al.}

\begin{abstract}
    Biomedical Question Answering aims to obtain an answer to the given question from the biomedical domain. Due to its high requirement of biomedical domain knowledge, it is difficult for the model to learn domain knowledge from limited training data. We propose a contextual embedding method that combines open-domain QA model \aoa and \biobert model pre-trained on biomedical domain data. We adopt unsupervised pre-training on large biomedical corpus and supervised fine-tuning on biomedical question answering dataset. Additionally, we adopt an MLP-based model weighting layer to automatically exploit the advantages of two models to provide the correct answer. The public dataset \biomrc constructed from PubMed corpus is used to evaluate our method. Experimental results show that our model outperforms state-of-the-art system by a large margin.
\end{abstract}


\maketitle

\section{Introduction}

Question answering is a classic task in Natural Language Processing, requiring a model to understand natural lang-uages. Cloze-style question answering problem has been a popular task because it is relatively easier to build cloze-style datasets. The cloze style question aims to select the best candidate answer regarding the specified context and fill in the blank of the question. Multiple cloze-style datasets have been published, such as CNN/Daily Mail \cite{hermannTeachingMachinesRead2015}, Children's Book Test \cite{hillGoldilocksPrincipleReading2016},  etc. Models based on neural networks are proposed, such as AS READER \cite{kadlecTextUnderstandingAttention2016}, CAS READER \cite{cuiConsensusAttentionbasedNeural2016}, \aoa \cite{cuiAttentionoverAttentionNeuralNetworks2017}and BERT \cite{devlinBERTPretrainingDeep2019}.

These models have achieved good performance on several datasets. However, they do not perform well when facing domain-oriented problems. The main reason is that domain-oriented questions require more background knowledge to give an answer, and a large dataset is needed to allow the models to learn the required domain knowledge.

We make improvements to the existing model, \aoa, and validate our results on the \biomrc dataset\cite{pappasBioMRCDatasetBiomedical2020}, the public biomedical dataset constructed from corpus from PubMed. We put forward the Contextual Word Embedding method and the MLP-based model weighting strategy for the biomedical question answering task. By combining the open-domain QA model and domain-oriented contextual word embedding, the proposed method outperforms state-of-the-art system on biomedical domain question answering significantly, setting up a new state-of-the-art system.

The main contributions of this paper are listed as follows:
\begin{itemize}
    \item Combining \biobert and \aoa, which can take full advantage of contextual word embedding model pre-trained on large domain corpus and mining semantic and contextual information to choose the best answer. In particular, multiple aggregation methods are adopted and evaluated.
    \item An MLP-based model weighting strategy is proposed, which can automatically learn the preferences and biases of different models and exploit the advantages of both models to provide the correct answer.
    \item Our method is evaluated on the \biomrc dataset, and the results show that it outperforms state-of-the-art system significantly. Our code is available at \url{https://github.com/leoleoasd/MLP-based-weighting}.
\end{itemize}

\section{Related Work}

The research of Question Answering has made rapid progress, which benefits from the publication of large-scale and high-quality datasets. Richadson et al. release MCTest \cite{richardsonMCTestChallengeDataset2013}, a multiple choice machine reading comprehension dataset that opened up research on statistical-based machine learning models. Hermann et al. release the CNN Daily Mail dataset \cite{hermannTeachingMachinesRead2015}, which includes over 1 million cloze-style data. More high-quality datasets have been released since then, such as SQuAD \cite{rajpurkarKnowWhatYou2018,rajpurkarSQuAD1000002016}, Facebook Children's Book Test \cite{hillGoldilocksPrincipleReading2016}, etc.

Models based on deep learning technologies significantly outperform traditional models in extracting context information. Hermann et al. \cite{hermannTeachingMachinesRead2015} propose an attention-based neural network and proves that the incorporation of attention mechanism is more effective than traditional, statistical-based baselines. Seo et al. propose the BiDAF \cite{seoBidirectionalAttentionFlow2017} model, which uses different levels of encoding for linguistic representation and uses a bidirectional attention flow mechanism to obtain the query-aware context representation. Kadlec et al. \cite{kadlecTextUnderstandingAttention2016} propose a simple model, the \emph{Attention Sum Reader} (AS Reader), which uses attention to directly pick answer. Fu et al. propose Ea-Reader \cite{fuEAReaderEnhance2019}, whose memory updating rule is able to maintain the understanding of document through read, update and write operations. Chen et al. propose McR\textsuperscript{2} \cite{chenMultichoiceRelationalReasoning2020}, which enables relational reasoning on candidates based on fusion representations of document, query and candidates. Fu et al. propose ATNet \cite{fuATNetAnsweringClozeStyle2019}, which utilities both intra-attention and inter-attention to answer close-style questions over documents. Cui et al. propose \emph{Attention-over-Attention Reader} ( AoA READER) \cite{cuiAttentionoverAttentionNeuralNetworks2017}, which puts another level of document-to-query attention on top of query-to-document attention, achieving state-of-the-art performance on multiple datasets.

In recent years, researchers have focused on combining the unsupervised pre-training on large corpus and supervised fine-tuning on the specific task. Vaswani et al. propose the Transformer \cite{vaswaniAttentionAllYou2017} model, which uses attention mechanism to replace the CNN and RNN parts of the traditional model to improve the model while speeding up the training process. Devlin et al. build a large-scale unsupervised pre-training model BERT \cite{devlinBERTPretrainingDeep2019} on top of Transformer to pre-train the language representation model using the masked language model task and the next sentence prediction task, and innovatively propose a training strategy that separates pre-training and fine-tuning. While sharing the same pre-trained weights, BERT achieves state-of-the-art performance on many different down\-stream tasks and datasets. Other BERT-based model is proposed, for instance, Liu et al. propose {\scshape RoBERTa}\cite{liuRoBERTaRobustlyOptimized2019}, providing servel techniques to robustly pre-train language models. Lan et al. propose ALBERT\cite{lanALBERTLiteBERT2019}, providing two parameter-reduction techniques to lower memory consumption and increase the training speed of BERT.

In the biomedical domain, Tsatsaronis et al. launch the BioASQ \cite{tsatsaronisOverviewBIOASQLargescale2015} challenges. It contains multiple subtasks, including article / snippet retrieval, document classification and question answering. Pappas et al. construct two cloze-style datasets, { \scshape bioread}\cite{pappasBioReadNewDataset2018} and \biomrc \cite{pappasBioMRCDatasetBiomedical2020}, and compare the accuracy of experts, non-expert human and different baseline models and neural network models. The results show that the baseline methods fail to correctly answer questions of the \biomrc dataset while neural MRC models perform well, indicating that the \biomrc dataset is less noisy and has enough features for the model to learn. Tang et al. initiate the CORD-19 \cite{tangRapidlyBootstrappingQuestion2020} dataset at the beginning of the global COVID-19 pandemic to help researchers in the biomedical field retrieve articles quickly.

\begin{figure*}[t]
    \centering
    \includegraphics[width=\linewidth]{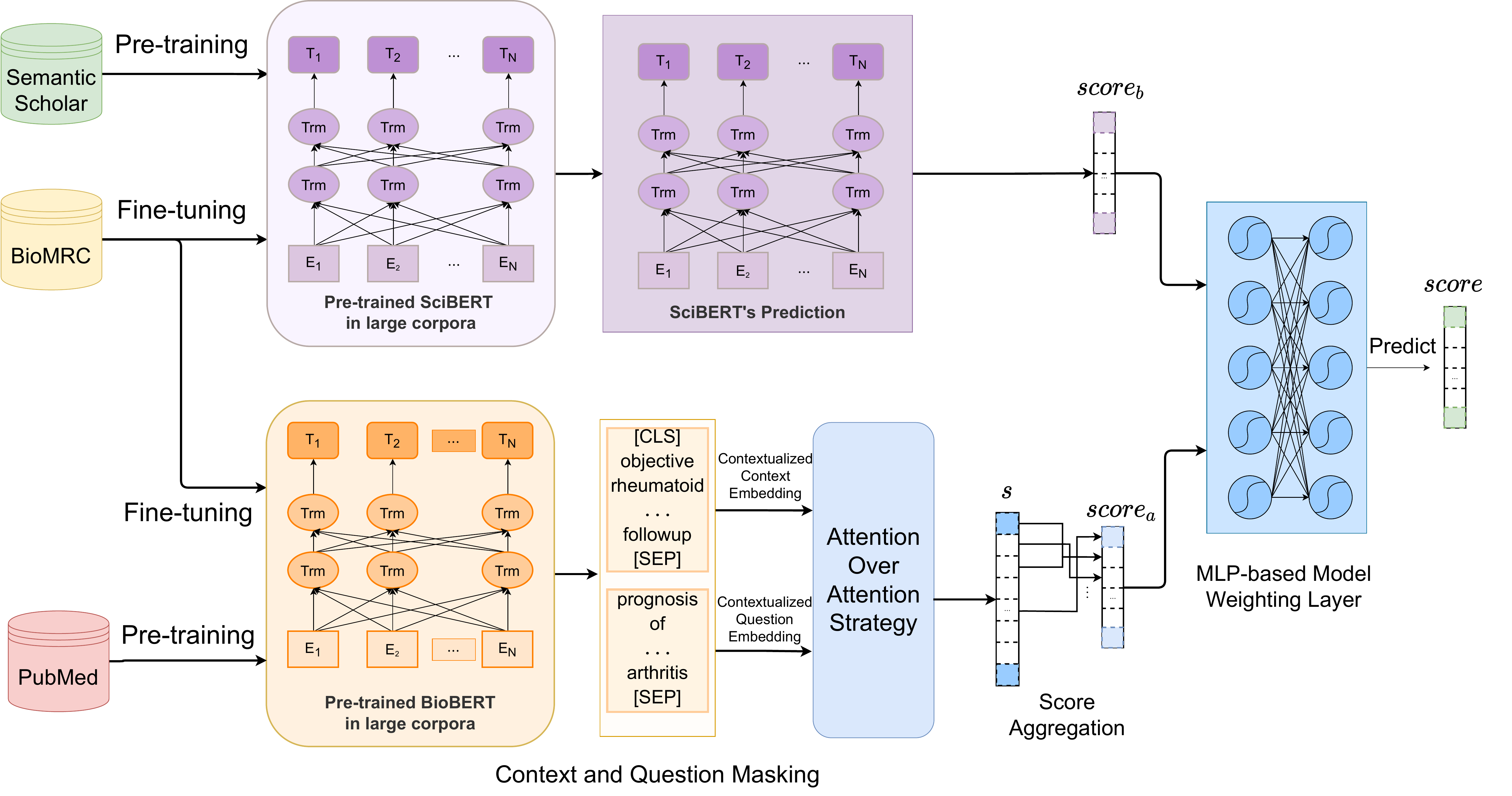}
    \caption{Model structure based on Pre-training and Weighting Strategy}
    \label{fig:overview}
\end{figure*}

Traditional neural network models require large-scale, high-quality supervised training data to obtain better results. However, it is challenging to build a large-scale and high-quality dataset for domain-oriented tasks because it requires domain experts' annotations. Recent researchers prove that combining the pre-training on large corpus and fine-tuning on supervised training data can achieve better performance on domain-oriented tasks. Gururangan et al. \cite{gururanganDonStopPretraining2020a} proves that a second phase of pre-training in domain (domain-adaptive pre-training) leads to performance gains. Gu et al. propose BLURB (Biomedical Language Understanding \& Reasoning Benckmark) \cite{guDomainSpecificLanguageModel2021} to test the biomedical language understanding ability of language models. Lee et al. propose BioBERT \cite{leeBioBERTPretrainedBiomedical2019} and Cohan et al. propose SciBERT \cite{beltagySciBERTPretrainedLanguage2019}, both of which learn the corpus representation of papers on a large-scale corpus of biomedical papers or scientific papers and have obtained better results on natural language processing tasks in biomedical domains. We aim to further improve the performance by using \biobert to obtain biomedical contextual and semantic information and by using model weighting layers to combine different models.

\section{Method}
We propose a pre-training strategy based on the scientific pre-training model (\scibert) and open-domain QA model (\aoa ) to obtain the final answer to the question. In particular, different embedding and weighting strategies are used in the training process. \cref{fig:overview} shows the full structure of our model.

\subsection{Formal Task Description}
This model is aiming at tasks that comprise cloze-style questions, to which answers are closely related to the comprehension of the context documents included in the problems. A set of candidate answers are also provided alongside, and the model is supposed to choose an answer from the candidates. This task can be formalized as a triplet $\left\langle \mathcal{C}, \mathcal{Q}, \mathcal{A} \right\rangle$ that is inclusive of the given context $\mathcal{C} = \set{w_1, w_2, \dots, w_n}$ made of words $w_i$, a query $\mathcal{Q} = \set{q_1, q_2, \dots, [MASK] , \dots, q_m}$ where the special token [MASK] marks the position where the answers are supposed to be placed, and answer candidates $\mathcal{A} = \set{a_1, a_2, \dots, a_o}$. A function $F$ is expected to be learned by the model to predict the answer $\mathcal{A}$ of question $\mathcal{Q}$ based on its comprehension of the proffered context $\mathcal{C}$:
\begin{gather}
    \forall a \in \mathcal{A}, P(a|\mathcal{C}, \mathcal{Q}) = \begin{cases}
        1 & a \text{ is the correct answer} \\
        0 & a \text{ is not the correct answer}
    \end{cases} \\
    F(\mathcal{C}, \mathcal{Q}, \mathcal{A}) = \max_{a \in \mathcal{A}} P(a|\mathcal{C}, \mathcal{Q})
\end{gather}

\subsection{Training of \scibert}

\subsubsection{Training data}

We use the \scibert \cite{beltagySciBERTPretrainedLanguage2019}, which has been pre-trained on the Semantic Scholar corpus. The corpus consists of 18\% papers from the computer science domain and 82\% from the broad biomedical domain. The unsupervised pre-training process using the large-scale corpus allows the model to obtain the semantic information of biomedical texts. Further, in order to let the model adapt to the cloze-style question answering task, \biomrc dataset is used to fine-tune the model.

\subsubsection{Answer extraction}

We adopt the answer extraction strategy Pappas et al. used in their \biomrc dataset \cite{pappasBioMRCDatasetBiomedical2020}. For each context-question pair, we first divide the context into sentences using NLTK\cite{birdNLTKNaturalLanguage2004}. Each sentence is concatenated to the question by [SEP] token, and they are fed to \scibert respectively. In this way, we obtain the top-level embedding of the candidate entities and the placeholder in the question. The embeddings of each entity in the sentence are connected to the placeholder's embedding and are sent to a multilayer perceptron to obtain the score for the particular entity. If an entity appears multiple times in the paragraph, we choose the maximum value of its score.

\subsection{Training of \aoa}

In order to make \aoa \cite{cuiAttentionoverAttentionNeuralNetworks2017} achieve better performance on domain-oriented tasks, we adopt different contextualized word embedding and attention aggregation strategies.

\subsubsection{Contextualized word embedding}

The \aoa uses direct word embedding. This approach converts each token in the context $\mathcal{C}$ and the question $\mathcal{Q}$ into a one-hot vector. It then converts them into a continuous representation using the shared embedding matrix $W_e$. Using this method in biomedical question answering may result in severe Out-Of-Vocabulary issues, given that most terms in the biomedical domain are made through several word-formation methods. Meanwhile, the model cannot learn necessary knowledge and the meanings of terms due to the lack of domain-oriented training data.

Instead, we adopt \biobert to generate contextualized word embedding, which has been pre-trained on a large biomedical corpus from \pubmed containing biomedical literature proposed by Devlin \cite{devlinBERTPretrainingDeep2019}. Thus, we can obtain domain-oriented knowledge and terms.

For tokenization, WordPiece \cite{wuGoogleNeuralMachine2016} through which new words can be represented by known tokens is used. Further, it can solve the Out-Of-Vocabulary issue and allow the model to better understand domain terms made by word-formation methods.

After connecting the context and the question with [SEP] token, it is fed into \biobert. If it is longer than length limit (512 tokens), we trim the back of the context and keep the original question unchanged:
\begin{equation}
    E(\mathcal{C, Q}) = BERT([\mathrm{[CLS]};\mathcal{C}; \mathrm{[SEP]}; \mathcal{Q};\mathrm{[SEP]}])
    \label{fig:bert}
\end{equation}
$\mathrm{E(\mathcal{C, Q})}$ in \cref{fig:bert} is the contextual embedding of the context $\mathcal{C}$ and the query $\mathcal{Q}$. 

To obtain the embeddings of context $\mathcal{C}$ and the query $\mathcal{Q}$, we've applied a masking operation on $E(\mathcal{C, Q})$ for segmentation. This would conceal the representation of the other segment by zero vectors, leaving the desired half acquired:
\begin{align}
    E(\mathcal{C})_i &= \begin{cases}
        E(\mathcal{C, Q})_i & i \text{ is a context token} \\
        0 & i \text{ is a question token} \\
    \end{cases} \\
    E(\mathcal{Q})_i &= \begin{cases}
        E(\mathcal{C, Q})_i & i \text{ is a question token} \\
        0 & i \text{ is a context token.} \\
    \end{cases}
    \label{eq:masking}
\end{align}

We adopt bi-directional RNN to further obtain the contextual representations $h_{context}(\mathcal{C}) \in \mathbb{R}^{(|\mathcal{C}| + |\mathcal{Q}| + 3) * 2d}$ of the context $\mathcal{C}$:
\begin{gather}
    \overrightarrow{h_{context}(\mathcal{C})} = \overrightarrow{GRU}(E(\mathcal{C})) \\
    \overleftarrow{h_{context}(\mathcal{C})} = \overleftarrow{GRU}(E(\mathcal{C})) \\
    h_{context}(\mathcal{C}) = [\overrightarrow{h_{context}(\mathcal{C})}; \overleftarrow{h_{context}(\mathcal{C})}]
\end{gather}
and similarly we obtain $h_{question}(\mathcal{Q}) \in \mathbb{R}^{(|\mathcal{C}| + |\mathcal{Q}| + 3) * 2d}$ for the question $\mathcal{Q}$.

\subsubsection{Pair-wise Matching Score}

After obtaining the contextual embedding of the context $h_{context}$ and the question $h_{question}$, we calculate a pairwise matching matrix, which indicates the relevance between a token in the context and question, by calculating their dot product:
\begin{equation}
    M(i, j) = h_{context}(i)^T \cdot h_{question}(j)\quad \; i \in \mathcal{C}, \; j \in \mathcal{Q}
\end{equation}

\subsubsection{Attentions over Attention Mechanism}

\begin{figure*}[t]
    \begin{tblr}{colspec=cX,hlines,vlines}
        Context & Because of reports of anaplastic transformation following irradiation, this study examines the incidence of anaplastic transformation and local control of these lesions. This review of seven \textcolor{cyan!80!white}{@entity1} who had \textcolor{orange!90!black}{@entity189} of the \textcolor{cyan!80!white}{@entity135} that was treated with irradiation shows local control in 71\% of cases. There were no cases of anaplastic transformation. This report adds to the literature two cases of "de-differentiation" to less differentiated \textcolor{cyan!80!white}{@entity957} ; one such case occurred after surgery alone. The literature is reviewed. Overall, anaplastic transformation is reported in 7\% of \textcolor{cyan!80!white}{@entity1} who had irradiation. De-differentiation occurs after surgery as well. The rate of local control with irradiation is less than 50\%; with surgery it is 85\%. It is concluded that surgery should be used if the procedure has acceptable morbidity. Otherwise, irradiation can be used. Failures can be salvaged surgically. "Anaplastic transformation" should not affect treatment approach. \\
        {Candidate\\Entities} & \textcolor{cyan!80!white}{@entity1}: [`patients']
        \textcolor{cyan!80!white}{@entity135}: [`head and neck']
        \textcolor{cyan!80!white}{@entity957}: [`squamous carcinomas']
        \textcolor{orange!90!black}{@entity189}: [`verrucous carcinoma'] \\
        Question & Radiotherapy in the treatment of \textcolor{magenta!50!purple}{XXXX} of the \textcolor{cyan!80!white}{@entity135} . \\
        Answer & \textcolor{orange!90!black}{@entity189}: [`verrucous carcinoma']
    \end{tblr}
    \caption{A example of the \biomrc dataset.}
    \label{fig:biomrc-example}
\end{figure*}

After getting the pair-wise matching matrix M, column-wise softmax is used to get the context-level attention regarding each token in the question:
\begin{gather}
    \alpha(t) = \softmax(M(1, t), \dots, M(|\mathcal{C}|, t)) \\
    \alpha = [\alpha(1), \alpha(2), \dots, \alpha(|\mathcal{Q}|)]
\end{gather}

We calculate a reversed attention using row-wise softmax to obtain the ``importance'' of each token in the question regarding each token in the context:
\begin{gather}
    \beta(t) = \softmax(M(t, 1), M(t, 2), \dots, M(t, |\mathcal{Q}|)).
\end{gather}

We average all $\beta(t)$ to get a question-level attention $\beta$:
\begin{gather}
    \beta = \frac{1}{n}\sum_{t=1}^{|\mathcal{C}|}\beta(t)
\end{gather}

Then we adopt the attention-over-attention mechanism, by merging these two attentions to get the ``attended context-level attention'':
\begin{gather}
    s = \alpha^T \cdot \beta
\end{gather}
where $s$ denotes the importance of each token in the context.

\subsubsection{Final Predictions}
\aoa regards an entity as a word no matter how many words it has. It uses \emph{sum attention} mechanism proposed by Kadlec et al. \cite{kadlecTextUnderstandingAttention2016} to get the confidence score of each candidate entity. In our model which uses WordPiece to obtain contextualized word embeddings, an entity may be either segmented into multiple tokens or composed of multiple words, and each token of the entity may occur multiple times in the context. So the confidence score of each candidate answer $a$ is calculated by aggregating all the occurrences of all its tokens in the context:
\begin{equation}
    P(a|\mathcal{C}, \mathcal{Q}) = \fa_{t \in \mathcal{T}(a)}(\fb_{i \in I(t, \mathcal{C})}(s_i))
\end{equation}
where $\mathcal{T}(a)$ is the result of segmenting the candidate answer $a$ using WordPiece; $F_1$ and $F_2$ are aggregating functions, which can be either \textbf{maximum} or \textbf{sum}, and $I(t, \mathcal{C})$ indicates the position that the token $t$ appears in the context $\mathcal{C}$.

\subsection{Model Weighting strategy}
After completing the training of \aoa and \scibert, a model weighting strategy is used to obtain the final answer by combining the advantages of both models.

Our previous study \cite{duDualModelWeighting2021} demonstrates that better performance can be achieved using a dual-model weighting strategy. The weighting process is performed by calculating a weighted average of the answer's confidence score and the similarity of the answer derived from two models. Further, considering that different models perform differently against data with different features, we use a simple MLP with one hidden layer to allow the model weighting layer to \emph{automatically} learn this difference and to take advantage of both models.
\begin{equation}
    score = MLP([score_a, score_b])
\end{equation}
Where $score_a, score_b \in \mathbb{R}^{|\mathcal{A}|}$ is the confidence score of each model.
In this way, the weighting layer would be able to learn the predilection and biases of each candidate model and achieve better performance.

\section{Experiment}
\subsection{Datasets}
We conduct the experiments on the \biomrclite dataset \cite{pappasBioMRCDatasetBiomedical2020} to verify the effect of our method. \biomrc is a biomedical cloze-style dataset for machine reading comprehension. The contexts in each sample are extracted from PUBTATOR, a repository containing 25 million abstracts and their corresponding titles on \pubmed. Biomedical entities in the abstract are extracted to form the candidate entities. The contexts are the abstracts themselves, and the questions are construct by randomly replacing a biomedical entity in the title with a placeholder. \cref{fig:biomrc-example} gives a sample in the \biomrc dataset.

\subsection{Experiment Settings}

\begin{table*}[t]
    \caption{{ \scshape The result of different aggregation functions, compared to the state-of-the-art model and human experts }}
    \label{tab:context-result}
    \begin{booktabs}{
        width=.8\linewidth,
        colspec={X[5] X[2.5] X[2.5] X[2.4] X[1.3] X[1.3] X[1.3]},
        cell{1}{1,2,3,4}={r=2,c=1}{},
        cell{1}{5}={r=1,c=2}{},
        cell{7}{1}={r=4,c=1}{},
        columns={valign=m,halign=c},
        column{1}={font=\scshape},
        column{4-7}={colsep=0pt},
        column{1-3}={colsep=2pt},
        cell{3-10}{2-7}={font=\normalfont},
        cell{10}{5-7}={font=\bfseries}
    }
        \toprule
        Method & Occurrence Aggregation & Token Aggregation & {{{Train Time\footnotemark[1]}}} & {{{\biomrclite}}} & & {{{\biomrctiny \footnotemark[2]}}} \\
        \cmidrule[lr]{5-6} \cmidrule[lr]{7}
         &  &  &  &  {{{Dev Acc}}} & {{{Test Acc}}} & {{{Test Acc}}} \\
        \midrule
        AS-READER                       & - & - & 16.56hr   & 62.29 & 62.38 & 66.67 \\
        AoA-READER                      & - & - & 60.90hr   & 70.00 & 69.87 & 70.00 \\
        SCIBERT-MAX-READER              & - & - & 83.22hr   & 80.06 & 79.97 & 90.00 \\
        Human Experts   & - & - & {{{-}}}         & {{{-}}} & {{{-}}} & 85.00 \\
        \midrule
        { AoA-READER with \biobert Embedding} 
                                        & max & max & 1.50hr &  78.54 & 78.11 & 90.00 \\
                                        & max & sum & 0.88hr &  83.40 & 83.36 & 93.33 \\
                                        & sum & max & 3.60hr &  80.98 & 81.20 & 90.00 \\
                                        & sum & sum & 1.76hr &  87.22 & 86.74 & 93.33 \\
        \bottomrule
    \end{booktabs}
    \vspace{1em}

    1: We conduct some code optimizations on the AoA-Reader model, so the training time of our implementation with BioBERT can not be compared to their original implementation.

    2: The test set of \biomrctiny dataset only contains 30 samples, and so the results on it may be unstable. On the other hand, the demonstrated accuracy of human experts comes from averaging the results of multiple experts, so it is a bit more stable than other results.
\end{table*}

Our experiments are carried out on the machine with Intel i9-10920X (24) @ 4.700GHz, GPU of GeForce GTX 3090 24G, using pytorch 1.9.1 as the deep learning framework. To avoid overfitting, all models are trained for a maximum of 40 epochs, using early stopping on the dev, with a patience of 3 epochs.

During the process of fine-tuning \scibert, the batch size is set to 1 and the top layer of \scibert is frozen; other layers are trained with the learning rate of 0.001.

During the process of fine-tuning \biobert and training \aoa, the batch size is set to 30, the learning rate is set to 0.001, and the learning rate for \biobert is set to $10^{-5}$. To reduce GPU memory usage, we use the mixed-precision training technique \cite{micikeviciusMixedPrecisionTraining2018}, setting precision to 16 bits.

We train our model on the \biomrclite dataset and evaluate it both on the \biomrc { \scshape lite } and { \scshape tiny } dataset, which have 100,000 and 30 samples, respectively. We use Setting A for \biomrc, in which all pseudo-identifier like \emph{@entity1} have a global scope, i.e., all biomedical entities have a unique pseudo-identifier in the whole dataset.

The model weighting layer is implemented after completing the training process of the two models, SciBERT and \aoa. The best weights evaluated by the \emph{Dev Acc} are chosen to obtain the individual scores of each sample, which will later be used to train the model weighting layer.

\subsection{Results}

\subsubsection{Performance of the Contextualized Word Embedding Strategy}

Contextualized word embedding strategy based on \biobert is used to obtain the final prediction answer. The selection of aggregating functions is crucial to the model performance. Therefore, multiple combinations of different aggregating functions are evaluated, and the results are shown in \cref{tab:context-result}.

\begin{table*}[htp]

    \caption{\textsc{The results of two single models and our MLP-based model weighting model, compared to the union accuracy of two single models.
    }}
    \label{tab:weighting-result}
    \begin{booktabs}{
        width=.8\linewidth,
        cell{1}{2}={r=1,c=2}{},
        cell{1}{1}={r=2,c=1}{},
        columns={valign=m,halign=c},
        column{1}={font=\scshape},
        colspec={X[4] X X X},
        row{3-5}={20pt},
        cell{5}{2-4}={font=\bfseries},
    }
        \toprule
        Method & {{{\biomrclite}}} & & {{{\biomrctiny}}} \\
        \cmidrule[lr]{2-3} \cmidrule[lr]{4}
        & {{{Dev Acc}}} & {{{Test Acc}}} & {{{Test Acc}}} \\
        \midrule
        AoA-READER with \biobert Embedding          & 87.22 & 86.74 & 93.33 \\
        SCIBERT-MAX-READER         & 79.74 & 80.21 & 86.67 \\
        MLP-based Weighting Model (ours)            & 88.76 & 88.00 & 96.66 \\
        \midrule
        The union of two single models (ideal result)       & 93.07 & 92.26 & 96.66 \\
        \bottomrule
    \end{booktabs}
    \vspace{1em}

\end{table*}

\begin{table}[t]
    \caption{{ \scshape The results of our MLP-based Weighting Model, excluding data that both models failed to answer }}
    \label{tab:weighting-result-best}
    \begin{booktabs}{
        width=\linewidth,
        cell{1}{2}={r=1,c=2}{},
        cell{1}{1}={r=2,c=1}{},
        columns={valign=m,halign=c},
        column{1}={font=\scshape},
        colspec={X[4] X X X},
        row{3-5}={20pt},
        cell{5}{2-4}={font=\bfseries},
    }
        \toprule
        Method & \biomrclite & & \biomrctiny \footnotemark[1] \\
        \cmidrule[lr]{2-3} \cmidrule[lr]{4}
        & {{{Dev Acc}}} & {{{Test Acc}}} & {{{Test Acc}}} \\
        \midrule
        AoA-READER with \biobert Embedding   & 93.71 & 93.21 & 96.55 \\
        SCIBERT-MAX-READER                         & 86.67 & 86.94 & 89.66 \\
        \midrule
        MLP-based Weighting Model (ours)            & 95.36 & 95.38 & 100.00 \\
        \bottomrule
    \end{booktabs}
    \vspace{1em}

    1: The test set of \biomrctiny dataset only contains 30 samples, and so the results on it may be unstable. 
\end{table}

It can be seen that choosing sum as both aggregation functions obtains better performance, and our model outperforms the state-of-the-art model significantly, which is about 6.77\% absolute improvements on the \biomrclite test sets.

Our model also shows an improvement on the \biomrctiny dataset, though the dataset contains only 30 samples, and this result may be unstable. Our performance on the larger \biomrclite test set still exceeds the average human expert performance on the \biomrctiny test set.

\subsubsection{Performance of the Weighting Model}

The MLP-based weighting model is used to further achieve better performance. We implement the SCIBERT-MAX-READER proposed by Pappas et al. \cite{pappasBioMRCDatasetBiomedical2020}, and use the model weighting strategy on top of \aoa with \biobert embedding and SCIBERT-MAX-READER. The results of our experiments are shown in \cref{tab:weighting-result}. The result of SCIBERT-MAX-READER comes from our implementation of this model, which is used to train our MLP-based weighting layer. The results slightly differs to those in \cref{tab:context-result}.

It can be seen that our MLP-based weighting model improves the accuracy effectively. Especially, the accuracy on the test set is improved by 1.26\% over the \aoa on the \biomrclite test dataset.

\begin{figure}[t]
    \begin{tikzpicture}    
        \begin{ganttchart}[
            canvas/.append style={draw=none,fill=none},
            today label font=\small\bfseries,
            title/.style={draw=none, fill=none},
            title label font=\bfseries,
            include title in canvas=false,
            bar label font=\mdseries\small\color{black},
            bar label node/.append style={left=.3cm},
            bar/.append style={draw=none, fill=blue!50!white},
            expand chart=\linewidth,
            y unit chart=.5cm,
            bar height=.8,
          ]{0}{100}
            \ganttbar[
                bar/.append style={fill=cyan6!30!cyan7},
                progress label text={6250},
                bar progress label anchor=west,
                progress=100]
            {All Questions}{0}{100} \\
            \ganttbar[
                bar/.append style={fill=cyan7},
                progress label text={5421},
                bar progress label anchor=west,
                progress=100
            ]{AoA-READER}{0}{87} \\
            \ganttbar[
                bar/.append style={fill=cyan7!60!cyan8},
                progress label text={5013},
                bar progress label anchor=west,
                progress=100
            ]{SciBERT}{12}{92} \\
            \ganttbar[
                bar/.append style={fill=cyan7!30!cyan8},
                progress label text={4668},
                bar progress label anchor=west,
                progress=100
            ]{Both}{12}{87} \\
            \ganttbar[
                bar/.append style={fill=cyan8},
                progress label text={484},
                bar progress label anchor=west,
                progress=100
            ]{Neither}{93}{100} \\
            \ganttbar[
                bar/.append style={fill=cyan8!50!cyan9},
                progress label text={5471},
                bar progress label anchor=west,
                progress=100
            ]{MLP weighting}{2}{90}
            \ganttbar[
                bar/.append style={fill=cyan8!50!cyan9},
                progress label text={26},
                progress=100
            ]{}{93}{93}
        \end{ganttchart}
    \end{tikzpicture}
    \caption{ The number of question answered correctly by different models on the \biomrclite dataset.}
    \label{fig:visualization}
    \Description{There were 6250 questions, and the \aoa correctly answered 5421 questions. SciBERT correctly answered 5013 questions, and there were 4668 questions that they both answered correctly. There were 484 questions that they both did not answer correctly, and the MLP weighted model could answer all of the questions that they both answered, most of the questions that one of them answered, and 26 questions that neither model answered.}
\end{figure}
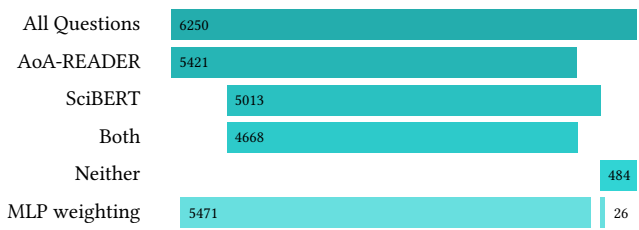

In order to evaluate the effectiveness of the model weighting layer, we compare the result to the union accuracy of two single models, i.e., the percentage of the union of the questions answered correctly by the two models in the total number of questions.

The results when excluding data that both models failed to answer are shown in \cref{tab:weighting-result-best}.The number of questions answered correctly by different models on the \biomrclite dataset is shown in \cref{fig:visualization}. Here, \emph{Both} in the figure refers to the the questions correctly answered by AoA-READER and SciBERT, and \emph{Neither} in the figure refers to questions that cannot be answered correctly by any model. 

As expected, both of the two models correctly answer some questions that the other model failed to answer. The proposed MLP-based weighting model not only gives the correct answer to the question that at least one model answers correctly, but also a small number of questions that both models fail to answer. 

To further corroborate the model's improvements in performance, we've applied the McNemar test to the results. Letting null hypothesis $h_0$ be our model has the same performance as SciBERT-MAX-READER, the alternate hypothesis $h_1$ would be there is a notable difference between the performance of our model and SciBERT-MAX-READER. The $h_1$ is accepted by the test where n=2, significance $\alpha$ = 0.025. Since our model has a lower error rate in all tests, it sufficiently supports the hypothesis that our model has significantly outperformed the SciBERT-MAX-READER.

In general, our MLP-based weighting model improves the performance by 2.17\% significantly compared to the original single model. These results demonstrate that the proposed method can automatically learn the biases and preferences and exploit the strengths of both models to achieve better performance.

\begin{figure}[t]
    \includegraphics[width=\linewidth]{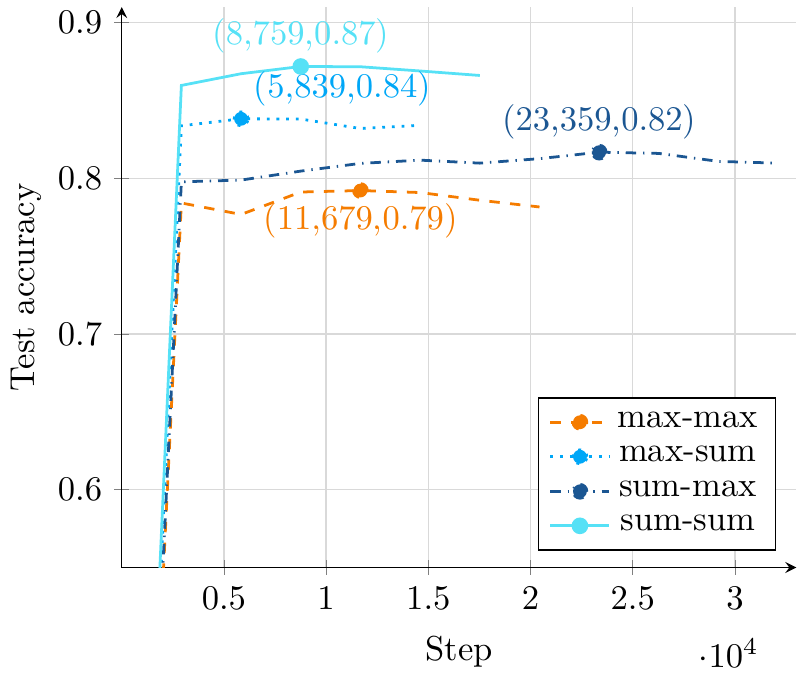}
    \caption{The validation accuracy of \aoa after each epoch.}
    \label{fig:loss}
    \Description{The valudation accuracy of \aoa with different token aggregation and occurence aggregation is shown.}
    Here, `max-max' represent token aggregation and occurence aggregation respectively.
\end{figure}

\subsubsection{Model Training Analysis}

In our structure, SCIBERT-MAX-READER and \aoa are trained separately, the outputs of which are then gathered for the final weighting layer to learn.

The \biobert embedded in the \aoa model is pre-trained on PubMed, and fine-tuned with the \aoa model on the BIOMRC dataset. It takes about 26 minutes for the \aoa to finish one epoch of learning based on the \biomrclite dataset.

All models generated in epochs are saved, among which the one that has the best performance on the dev set will be used to train the weighting layer.

The training process for the AoA reader is illustrated in \cref{fig:loss}. It can be seen that model using sum as both token and occurence aggregation converges faster compated to most models while giving best results.

\section{Conclusions}

We propose a contextual embedding and model weighting method, which can combine model pre-trained on a large corpus and open-domain QA model to mine semantic and contextual information in biomedical question answering. Especially, we adopt an MLP-based model weighting strategy which can automatically learn and utilize the preferences and biases of two models to combine their advantages. The results show that our method outperforms state-of-the-art system and has higher accuracy than experts. In future work, how to use the semantic similarity between entity tokens and context tokens in getting final predictions should be studied, i.e., a context token should contribute to the score of an entity if its semantic information is similar to that of entity token.

\bibliographystyle{ACM-Reference-Format}
\bibliography{references}

\end{document}